\documentclass[10pt,twocolumn,letterpaper]{article}

\usepackage{wacv}
\usepackage{times}
\usepackage{epsfig}
\usepackage{graphicx}
\usepackage{amsmath}
\usepackage{amssymb}
\usepackage{algorithm}
\usepackage[noend]{algpseudocode}

\wacvfinalcopy 

\def\wacvPaperID{575} 

\ifwacvfinal\fi
\begin{document}

\title{A Semi-Supervised Two-Stage Approach to Learning from Noisy Labels}

\author{Yifan Ding$^\dagger$ \hspace{2cm} Liqiang Wang$^\dagger$ \hspace{2cm} Deliang Fan$^\ddagger$ \\
$^\dagger$Department of Computer Science, University of Central Florida\\
$^\ddagger$Department of Electrical Engineering, University of Central Florida\\
{\tt\small yf.ding@knights.ucf.edu, lwang@cs.ucf.edu, dfan@ucf.edu}
\and
Boqing Gong \\
Tencent AI Lab\\
Bellevue, WA 98004\\
{\tt\small boqinggo@outlook.com}
}

\maketitle
\ifwacvfinal\thispagestyle{empty}\fi

\begin{abstract}

The recent success of deep neural networks is powered in part by large-scale well-labeled training data. However, it is a daunting task to laboriously annotate an ImageNet-like dateset. On the contrary, it is fairly convenient, fast, and cheap to collect training images from the Web along with their noisy labels. This signifies the need of alternative approaches to training deep neural networks using such noisy labels. Existing methods tackling this problem either try to identify and correct the wrong labels or reweigh the data terms in the loss function according to the inferred noisy rates. Both strategies inevitably incur errors for some of the data points. In this paper, we contend that it is actually better to ignore the labels of some of the data points than to keep them if the labels are incorrect, especially when the noisy rate is high. After all, the wrong labels could mislead a neural network to a bad local optimum. We suggest a two-stage framework for the learning from noisy labels. In the first stage, we identify a small portion of images from the noisy training set of which the labels are correct with a high probability. The noisy labels of the other images are ignored. In the second stage, we train a deep neural network in a semi-supervised manner. This framework effectively takes advantage of the whole training set and yet only a portion of its labels that are most likely correct. Experiments on three datasets verify the effectiveness of our approach especially when the noisy rate is high. 

\end{abstract}


\section{Introduction}
With the recent development of deep neural networks,  we have witnessed great advancements in visual recognition tasks such as image classification~\cite{russakovsky2015imagenet,szegedy2015going,sanchez2013image,simonyan2014very}, object detection~\cite{everingham2010pascal,uijlings2013selective,sermanet2013overfeat,girshick2014rich}, and semantic segmentation~\cite{long2015fully,ciresan2012deep,hariharan2014simultaneous,farabet2013learning}. Take the famed object recognition challenge ILSVRC~\cite{russakovsky2015imagenet} for instance, the Inception-ResNet-v2~\cite{szegedy2017inception} achieves a remarkable top-5 accuracy of 95.3\% in 2017. The success of the deep neural networks is powered in part by large-scale well-labeled training data. However, it is actually a very daunting task to laboriously annotate an ImageNet-like training set.

On the contrary, it is fairly convenient, fast, and cheap to collect training images from the Web along with their noisy labels. This signifies the need of alternative approaches to train deep neural networks using such noisy labels. Indeed, the very first source of training data is often from the Web when we face a new visual recognition task. Therefore, the methods that  effectively learn from the noisy  labels can significantly reduce the human labeling efforts, even to zero effort in some scenarios.

\begin{figure}
\centering
\includegraphics[width=0.5\textwidth]{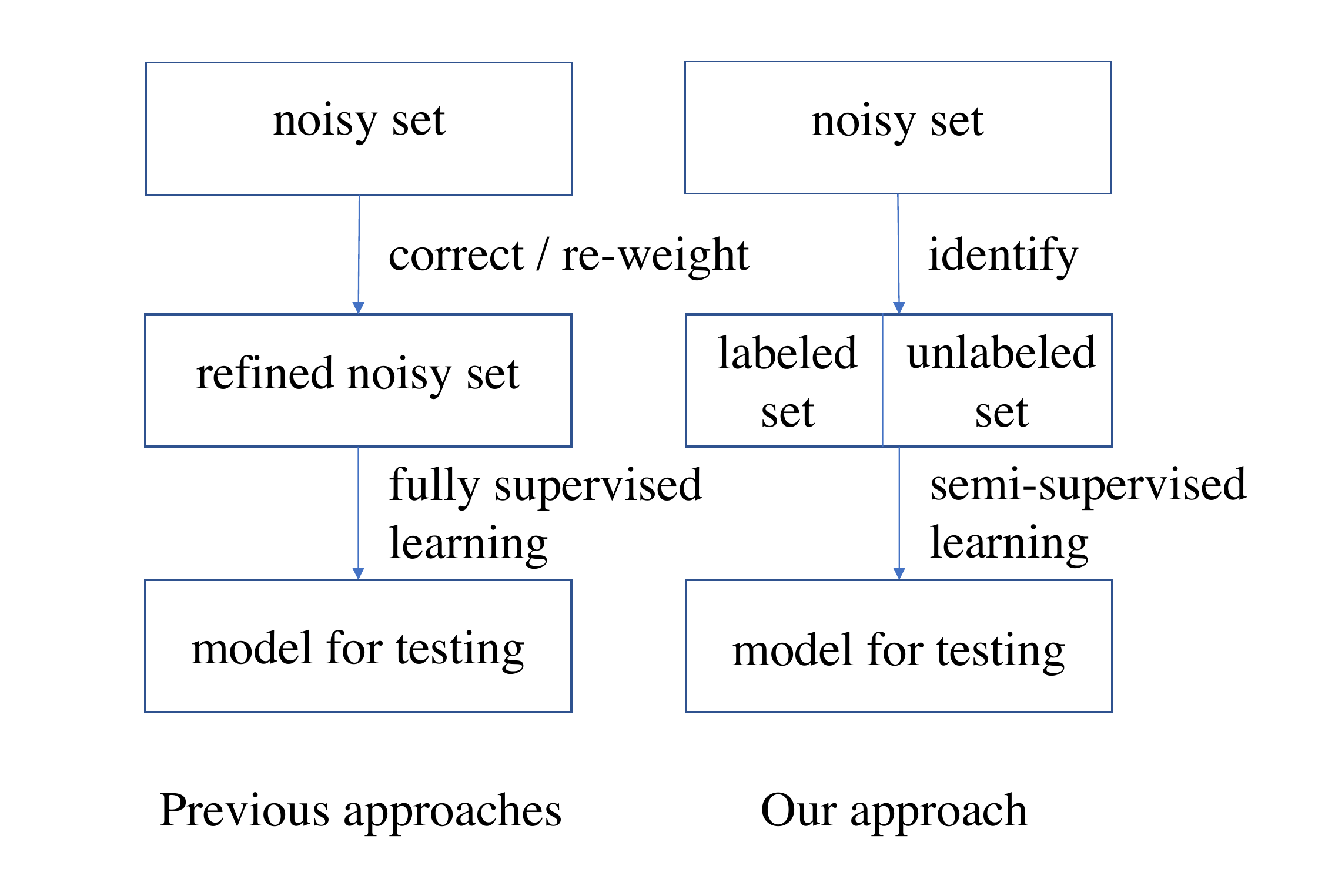}
   \caption{Comparison between our framework and the one of existing methods for learning from noisy labels. Instead of trying to correct or reweigh all the labels of the noisy data, we ignore the ambiguous ones and train a deep neural network in a semi-supervised fashion. }\label{fig:dif}
   \vspace{-5pt}
\end{figure}

There has been a rich line of recent works that aim to address the problem of learning from noisy labels. We categorize them to two groups: one directly learns from the noisy labels and the other relies on an extra set of clean data. For the former, a label cleansing module is often applied in order to identify the correctly labeled data~\cite{chen2015webly,gan2016webly,brodley1999identifying}. Alternatively, one may model the noise to reweigh the data terms in the loss functions~\cite{liu2016classification,patrini2016loss,natarajan2013learning}. The performance of these algorithms heavily depend on the precision of the label cleansing or the estimated noisy rates. They perform well when the noise rates can be safely managed~\cite{manwani2013noise,stempfel2009learning,natarajan2013learning,patrini2016loss}, but could suffer from the ambiguity between mislabeled examples and ``hard cases'' --- the data points whose labels are correct but hard to be captured by the neural networks' classification boundary. 

For the second group of methods, an extra set of clean data is used to guide the learning agent through the noisy data. Li et al.~\cite{li2017learning} enforce the network trained from the noisy data to imitate the behavior of another network learned from the clean set. Vahdat~\cite{vahdat2017toward} constructs an undirected graphical model to represent the relationship between the clean and noisy data. Veit et al.~\cite{veit2017learning} also use a secondary network to clean the labels of the noisy data such that the main network receives more accurate supervision than from the original training set. An absence of the clean set would prevent these methods from being applied to some situations. Besides, like the approaches in the first group, they still aim to correct the labels of the noisy set and could make mistakes in this procedure.



Despite their promising results, both groups of the existing methods come with a common caveat --- they attempt to correct the noisy labels or reweigh the terms of all the data points no matter how difficult it is to do so. This inevitably incurs errors for some of the data points. In this paper, we contend that it is better to completely ignore the labels of some of the data points than to keep their wrong labels. After all, the wrong labels could mislead the training procedure of the network. We suggest a two-stage framework for the learning from noisy labels. In the first stage, we identify a small portion of data points from the noisy training set of which the labels are correct with a high probability. The noisy labels of the other data points are then removed. In the second stage, we train a deep neural network in a semi-supervised manner. This framework effectively takes advantage of the whole training set and yet only a portion of its labels which is most likely correct. Figure~\ref{fig:dif} contrasts our framework to the one taken by most existing methods.




It is worth noting that the first stage of our framework can be implemented in a variety of ways. Many existing methods for learning from the noisy labels are actually applicable. This paper presents our preliminary study by a simple and efficient self-refining method for the first stage and leaves the exploration of more sophisticated approaches to the future work. In particular, we rank all the data points within each class and then keep the labels of the top few. The ranking is performed by the multi-way classification neural network learned from the original training set when there is no clean set available, and by a binary classifier of each class which is trained to differentiate the data of clean and noisy labels when the clean set is given. In the second stage, we apply the temporal ensembling~\cite{laine2016temporal} to train a deep neural network in the semi-supervised manner. To the best of our knowledge, this is the first time that the temporal ensembling is tested on a large set of natural images --- around 1M images of which the resolutions are about 256x256.

It is also worth noting that when there exists a small clean set in addition to the noisy training set, the semi-supervised learning approach~\cite{lee2013pseudo} has been considered a baseline in the experiments of~\cite{xiao2015learning}. In a sharp contrast to the observations of~\cite{patrini2016making}, we show that, under our two-stage framework, semi-supervised learning methods can actually give rise to state-of-the-art results for the task of learning from noisy labels.


The rest of this paper is organized as follows. Section~2 discusses several related areas to our method. In Section~3, our two-stage approach is described in details. Section~4 shows the experimental results of our approach in two diminutive datasets and a large-scale real noisy dataset. Finally, we conclude the paper in Section~5.

\section{Related work}

Our approach is broadly related to four research topics: learning from noisy labels, the robustness of neural networks, semi-supervised learning, and Webly-supervised learning. We discuss each of them as below.

\emph{Learning from noisy labels}: The purpose of learning from noisy labels is to deal with noisy labels in the training data and reduce its negative influence toward the accuracy of classifiers. Following the review in~\cite{frenay2014classification}, the algorithms of learning from noisy labels can be grouped into three clusters: noise-robust approaches~\cite{teng2000evaluating,teng2001comparison} which depend on the robustness of neural networks and do not really deal with noise, label noise-tolerant methods which usually make use of some side information like the noisy rate in each class to design models that account for the label noise~\cite{joseph1995bayesian,patrini2016making,joulin2016learning,li2017learning,reed2014training}, and label noise cleansing methods. In the third category, different approaches are proposed to either remove or correct the noisy labels.~\cite{sun2007identifying,aha1991instance} identify the mislabeled sample and reassign correct labels to them while~\cite{john1995robust,muhlenbach2004identifying}  delete possibly noisy samples. \cite{miranda2009use} removes and meanwhile corrects noisy labels of bio-informatics data sets. \cite{sukhbaatar2014training,xiao2015learning}  add an extra noisy layer to match the neural network outputs with the noisy label distribution. \cite{ramaswamy2016mixture} proposes a noisy estimator using kernel mean embedding. 


\emph{The robustness of neural networks}: it is worth mentioning that the stochastic nature of the training algorithms of neural networks tolerates noisy labels by itself to some extent. \cite{rolnick2017deep} shows that deep neural networks are able to learn from the majority clean data while the gradient updates from noisy samples cancel out in each batch of training samples.~\cite{sukhbaatar2014training} proves that a standard Convnet model ~\cite{krizhevsky2012imagenet} is surprisingly robust to label noise.~\cite{van2015building} also finds that learning algorithms based on CNN features and part localization are robust to mislabeled training examples when the error rate is not too high. This ability makes it possible to refine the noisy set or correct some labels by the network directly learned from the noisy labels.


\emph{Semi-supervised learning methods}: Semi-supervised methods are proposed to learn in the presence of both labeled and unlabeled data~\cite{zhu2009introduction}, which is naturally applicable to the problems of learning from noisy labels when there is an extra clean set, e.g., by concealing the labels of the noisy set. Ladder network~\cite{rasmus2015semi} is one of the methods that introduces lateral connections into an encoder-decoder network and it is trained to simultaneously minimize the sum of supervised and unsupervised cost functions by back-propagation in a layer-wise manner. Our work is mostly related to~\cite{laine2016temporal} which proposes a simpler but more efficient $\Pi$ model and a temporal model which only minimizes the difference of two predicted probabilities of the same inputs accumulated in different epochs. More recent semi-supervised learning methods~\cite{miyato2017virtual,tarvainen2017weight,odena2016semi} use adversarial samples to regularize the networks. Some graph based methods propagate labels among the training data~\cite{zhu2005semi,fergus2009semi}.  Intermediate predictions by the model under training are used as pseudo labels~\cite{lee2013pseudo} to reinforce the model. \cite{breve2010semi} uses a graph and labeled samples to push away mislabeled ones.

\emph{Webly-supervised learning methods}: Webly-supervised learning methods specify the training data gathered from the web~\cite{xiao2015learning,zhuang2016attend,gan2016webly,li2017webvision}. Usually, the gathered web data are companied with noise and have a very large scale; therefore, it is commonly impossible to know the exact noisy rate of each class. Under such a condition, estimation using a small portion of clean data~\cite{patrini2016making} or noise modeling methods~\cite{lee2013pseudo,xiao2015learning} are often applied. While in our approach, we refrain from modeling the label noise and use semi-supervised learning instead to automatically propagate the labels of the mostly correct ones to the remaining training data points.

\begin{figure*}
\begin{center}
\includegraphics[width=1\textwidth]{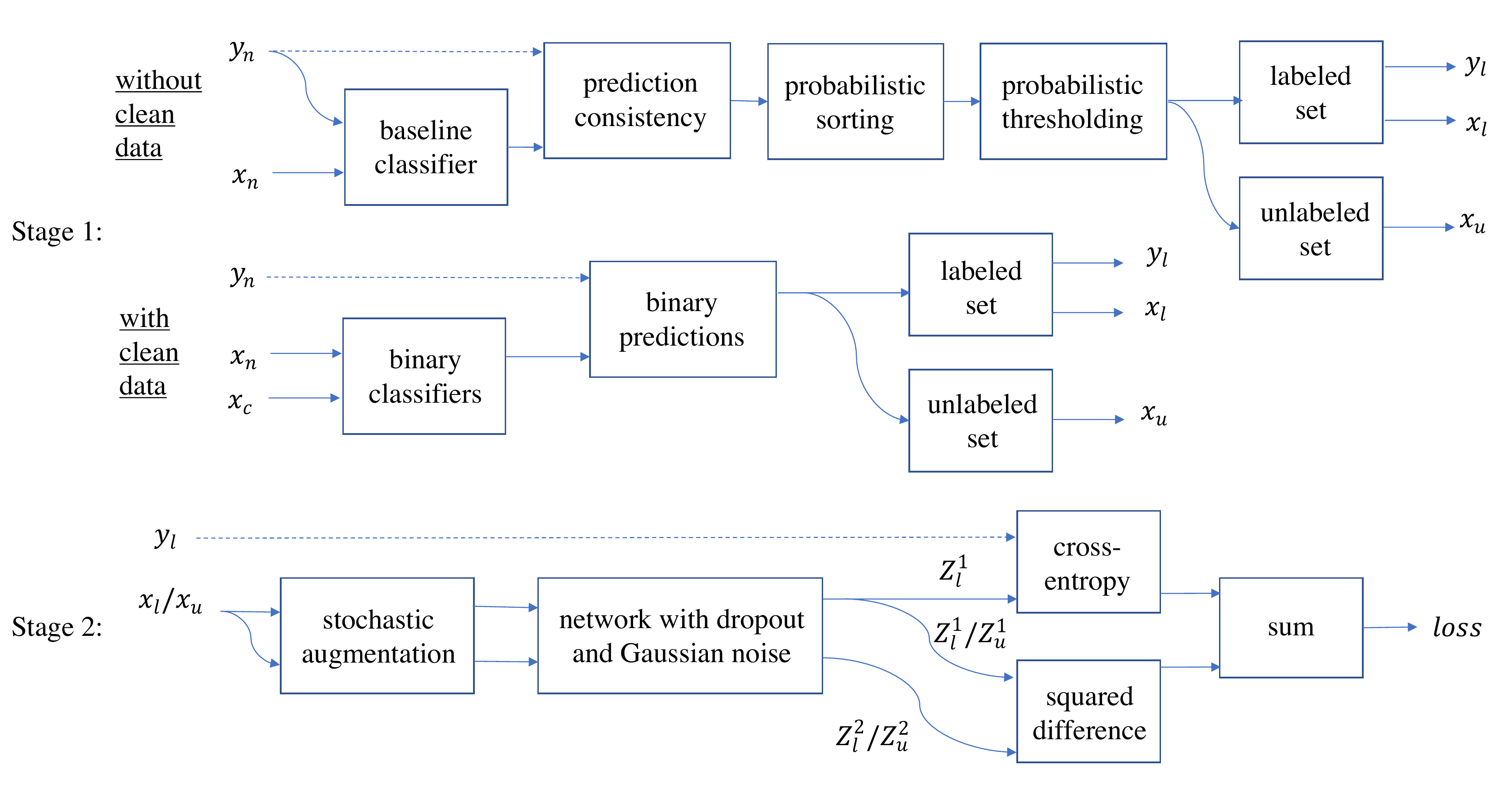}
\end{center}
\vspace{-5pt}
   \caption{Overview of the two stages in our approach. In Stage 1, we mine some examples from the noisy training set such that their labels are more likely correct than the others'. After that, we employ a semi-supervised learning method in Stage 2 to train a deep neural network. We use the subscripts $n$ and $c$ to represent the noisy and clean data, respectively. The subscripts $l$ and $u$ respectively stand for the labeled and unlabeled data points that are identified by Stage 1 and then input to Stage 2. }\label{fig:overview}
   \vspace{5pt}
\end{figure*}

\section{Approach}
We describe our  semi-supervised two-stage approach to learning from noisy labels in this section. It consists of two main components. In the first stage, we identify some data points from the noisy training set for which there exist strong indications that their labels are correct. In the second stage, a deep neural network is trained in a semi-supervised learning fashion over all the data points and yet using only the labels selected in the first stage. As a result, the network can be readily applied to classify previously unseen test data. As below we detail the implementations of the two components in this paper, and we stress that other realizations of the two-stage framework (cf.\ Figure~\ref{fig:dif}) can be explored in the future.

\subsection{Stage 1: identifying likely correct labels}
The goal of this stage is to mine the data points for which the labels are likely correct from the noisy training set. The selected data and their labels will be the seed for the semi-supervised learning approach in the second stage. Additionally, it is important to note that we have to also construct a good validation set in order to monitor the training procedure and choose proper hyper-parameters in the next stage.

We pre-train a deep neural network classifier using the noisy data. The trained neural network is then used to mine the seed data points for the second stage. The first two panels of Figure~\ref{fig:overview} illustrate this procedure.


%


In particular, we examine the data points class by class. First of all, we remove the labels of the data points for which the neural networks' predictions differ from their original noisy labels. This corresponds to the prediction consistency module in the first panel of Figure~\ref{fig:overview}. For the remaining images $\{x_i\}$ of a particular class $c$, we then rank them non-increasingly using the network's prediction $P(c|x_i)$. We keep the images for which the predicted probabilities are as high as 0.9. If less than 10\% of the images of this class are kept after that, we add more images down the ranking list such that the labels of the top 10\% of the images of that class pass the screening. We have also tested other percentages (e.g., 5\%) and do not observe any significant change of the experimental results. Table ~\ref{tab:ratio} shows the results (accuracies) on CIFAR10 when we keep the labels of four different percentages of the training set.

\begin{table}[h]
\caption{Performance of different percentages of labeled data points of the training set. (CIFAR-10, see Section 4 for detailed experimental setups).}\label{tab:ratio}
\begin{center}
\vspace{-10pt}
\begin{tabular}{c c c c c}
\hline
labeled \%& $p:0$ & sy.$p: 0.2$&asy.$p: 0.2$  & asy.$p: 0.6$ \\
\hline\hline
5\% & 87.9 & 84.2 & 85.5 &75.6\\
10\% & 88.0 & 84.5&85.6 &75.8\\
20\%& 87.9 & 85.1&85.5&76.2 \\  
50\%& 88.1 & 84.5 &85.6 &74.8 \\
\hline
\end{tabular}
\end{center}
\end{table}

{When an extra clean set is available,} more advanced techniques can actually be applied to mine the noisy set, e.g., using a graph between the clean and noisy sets~\cite{vahdat2017toward} or a secondary neural network trained from the clean set~\cite{veit2017learning}. We take an easy alternative and find it works well in the experiments. Specifically, we train a binary classifier for each class by assigning positive labels to the clean data of the class and negative labels to the data of other classes. After that, the classifier is used to classify the noisy data points of the corresponding class. We remove the labels of the images which are classified to the negative class. The second panel of Figure~\ref{fig:overview} demonstrates this procedure.


\begin{algorithm*}
\caption{The semi-supervised learning method used in Stage 2 of our method}\label{alg:semi}
\begin{algorithmic}[1]
\State $\textbf{Require: } x_i = \text{training stimuli}$
\State $\textbf{Require: } y_i = \text{labels for labeled data}$
\State $\textbf{Require: } |N| = \text{number of samples in one mini batch}$
\State $\textbf{Require: } |N/2| = \text{number of labeled data in one mini batch}$
\State $\textbf{Require: } f_\theta(x) = \text{stochastic neural network with trainable parameters }  \theta$
\State $\textbf{Require: } g(x) = \text{stochastic input augmentation function}$
\For {$e  \text{ in [ 1, numepochs]} $}
\For {$  \textit{each mini batch }  B $}

\State $Z_i^1 \in B \gets f_\theta(g(x_i \in B)) \text{  evaluate network outputs for augmented inputs}$
\State $Z_i^2 \in B \gets f_\theta(g(x_i \in B)) \text{  evaluate again the same inputs}$
\State $loss \gets - \frac{1}{N/2}\sum_{X_i \in X_l} \log Z_i[y_i] \text{  supervised loss component}$
\State $+ \frac{1}{N} \sum_{i=1}^N(||Z_i^1-Z_i^2||_2)^2 \text{ unsupervised loss component}$
\State update $\theta$ using, e.g., SGD  update network parameters
\EndFor
\EndFor
\State return $\theta$

\end{algorithmic}
\end{algorithm*}

\subsection{Stage 2: semi-supervised learning}
Recall that we aim to lean a good classifier neural network that can perform well at the test stage, not to correct the labels of the training set at all. In Stage 2, we train the classifier in a semi-supervised way using all the data points of the training set and yet keeping only the labels identified in Stage 1.

In particular, we use the methods proposed by~\cite{laine2016temporal} in our second stage. We improve the implementation of its $\Pi$ model for the learning from large-scale natural images~\cite{xiao2015learning} whose labels are noisy, thanks to its efficiency. For the other two smaller datasets, we directly borrow the original temporal ensembling model from~\cite{laine2016temporal} whose performance is superior over the simplified $\Pi$ model and yet has to be trained for many more epochs. We elaborate the $\Pi$ method in this section and refer the readers to \cite{laine2016temporal} for the details of the temporal ensembeling.

The main idea of the $\Pi$ method is to regularize the network such that it generates about the same outputs for the same input image that undergoes data augmentation and/or dropout twice. This is a reasonable regularization because the image labels are supposed to remain the same no matter how one augments the images. 


We explain the main idea using the third panel of Figure~\ref{fig:overview} and Algorithm~\ref{alg:semi}. Note that the subscripts are used to differentiate the labeled data points $x_l$ and those with no labels $x_u$ in the figure. Given an input image, no matter it is labeled or not, we apply some simple augmentations like horizontal flip and horizontal and vertical shift, and also add Gaussian noise. We further perturb the convolutional layers of the network by dropout. As a result, the same image $x_i$ actually incurs different output vectors $z_i^1$ and $z_i^2$ by the softmax layer of the network. Accordingly, a consistency regularization can be defined to push them close to each other,
\begin{equation} 
R = 1/N \sum_{i=1}^{N}||z_i^1 - z_i^2||_2^2
\end{equation}
where $N$ denotes the number of images in a mini-batch. 

Therefore, the overall cost function for training the network is the following,
\begin{equation} 
L = - 1/M \sum_{j=1}^M\log z_j[y_j] + \alpha /N \sum_{i=1}^{N}||z_i^1 - z_i^2||_2^2
\end{equation}
where the first term is the conventional cross-entropy loss over the $M$ labeled data points in a mini-batch. The notation $z_j[y_j]$ is to index the $y_j$-th element in the output vector $z_j$ and $y_j$ is the label of the data point $x_j$.

Since there are far more unlabeled data points than the labeled ones after the pruning of Stage 1, we sample the labeled data more frequently than the unlabeled in order to provide the network effective gradients. For each mini-batch of size $N$, we randomly choose $N/2$ labeled images and $N/2$ unlabeled ones.


In the second stage, our semi-supervised learning of the network continues from the one trained in Stage 1. We find that the values of the cross-entropy term are in general much larger than the regularization term. Therefore, the balance cost $\alpha$ is introduced and its value is determined according to the model's performance on the validation set. Unlike the ramp-up and ramp-down balance cost used in~\cite{laine2016temporal}, we fix $\alpha$ in the whole course of training for the ease of tuning on the large dataset. However, the ramp-up and ramp-down cost are used on the smaller MNIST and CIFAR-10 datasets.


\subsection{Balancing different classes in the training}

It is worth mentioning that we balance different classes in the training in our experiments because this simple and well-known trick gives rise to surprisingly large gains. By overly sampling the classes which have smaller number of samples than the other classes,  we make the training set balanced across all classes. We notice that for the images crawled from the Web, the long-tailed distribution of different classes is a common case. One can easily retrieve thousands of images of a popular query and yet only a few for less interesting queries or rare classes. This inevitably influences even the well benchmarked datasets. For example, for the classes in ImageNet~\cite{russakovsky2015imagenet}, a search of the keyword ``Goose'' returns a lot of results (and some of them are noisy). If we use the keyword ``Egretta Garzetta'', the results are very clean but there is only a small number of images. 

\section{Experiments}
We run extensive experiments to evaluate the proposed two-stage and semi-supervised approach on both small-scale MNIST~\cite{lecun1998gradient} and CIFAR-10~\cite{krizhevsky2009learning} and a large-scale benchmark of natural images, Cloting1M~\cite{xiao2015learning}. The results indicate that the two-stage method significantly outperforms the competing baselines when the noisy rate is high, and is comparable to the existing methods when the training set is contaminated by small noisy rates or zero noise.

\subsection{Datasets}

\newcommand{\specialcell}[2][c]{%
  \begin{tabular}[#1]{@{}c@{}}#2\end{tabular}}

We test our method on three datasets: MNIST~\cite{lecun1998gradient}, CIFAR-10~\cite{krizhevsky2009learning}, and Clothing1M~\cite{xiao2015learning}. MNIST is a dataset of handwritten digits. It has 60,000 training images and a test set of 10,000 image.  Each image has 28x28 pixels. CIFAR-10 consists of 60,000 32x32 tiny images of real objects. Among them, 10,000 are left out as the test set. The Clothing1M dataset is about the same scale as ImageNet but with only 14 clothing classes. According to the estimation by the authors, about 39\% of the labels in Clothing1M are incorrect because the dataset is automatically crawled by computer and has not been fully screened by human annotators. Nonetheless, for the research purpose, it does offer a clean test set of 22K images and a small extra clean set of 50K images that can be used in the training phase.

\begin{figure}
\begin{center}
\includegraphics[width=0.4\textwidth]{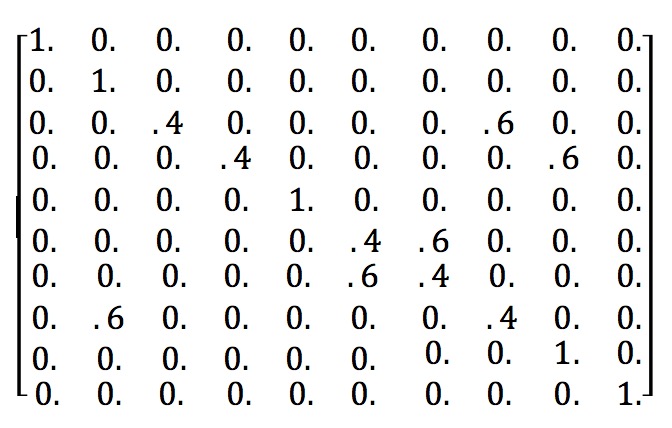}
\end{center}
\vspace{-5pt}
   \caption{An example of the label transition matrix for artificially adding noise to CIFAR-10.}\label{fig:t}
\end{figure}

We follow the experiment setups in~\cite{patrini2016making} to experiment with the three datasets. In particular, we add noise to the labels of the MNIST in the following way. Let $A\rightarrow_p B$ denote that the label of class A is changed to class B with probability $p$ for any data point of class A (for simplicity, $p$ is omitted in the discussion below). This artificially creates a noisy version of the MNIST. We change the labels following the paths $2 \rightarrow 7$, $3 \rightarrow 8$, $5 \rightarrow 6$, \text{and } $7 \rightarrow 1$. Similarly, we add noise to the labels of CIFAR-10 by TRUCK \( \rightarrow\) AUTOMOBILE, BIRD \( \rightarrow\) AIRPLANE, DEER \( \rightarrow\) HORSE, CAT \( \leftrightarrow\) DOG. When the noisy rate is $p=0.6$, an example transition matrix of the labels is shown in Figure~\ref{fig:t} for CIFAR-10. Both symmetric and asymmetric noises are tested in~\cite{patrini2016making}, and we  experiment with both as well.

For the images in the CIFAR-10  and Clothing1M datasets, we subtract the per-pixel mean from each of them before sending them to the network. In the training, we flip the images and randomly crop  32x32 and 224x224 regions from the CIFAR10 and Clothing1M images, respectively.

\subsection{Experiments on CIFAR-10 and MINIST}

For all the experiments on CIFAR-10 and MNIST, we leave 10\% out of the training set for validation and use the rest to train our neural networks. During the training, we decrease the learning rate by the ratio of 0.5 after the validation accuracy saturates for up to 10 epochs. We use the SGD optimizer for CIFAR-10 and AdaGrad for MNIST. The momentum is set to be 0.9, the initial learning rate is 0.1, and the minimal learning rate is set to be 1e-7. For all the baseline models, we use the cross-entropy loss as the objective function. We train our neural network for 40 epochs on MNIST and 120 epochs on CIFAR-10. 

For CIFAR-10, we use a 14-layers ResNet for both the baseline model and the semi-supervised learning model. Specifically for the semi-supervised learning model, we add a Gaussian noise layer on the top of the model and dropout between two convolutional layers within the residual units to achieve the output difference for the unsupervised loss. For MNIST, we implement a fully connected network with two dense hidden layers of size 128 and dropout. Again, we add Gaussian noise layer for the semi-supervised learning model. Both networks follow the same architectures as in~\cite{patrini2016making}.


\begin{figure*}[h]
\begin{center}
\includegraphics[width=1\textwidth]{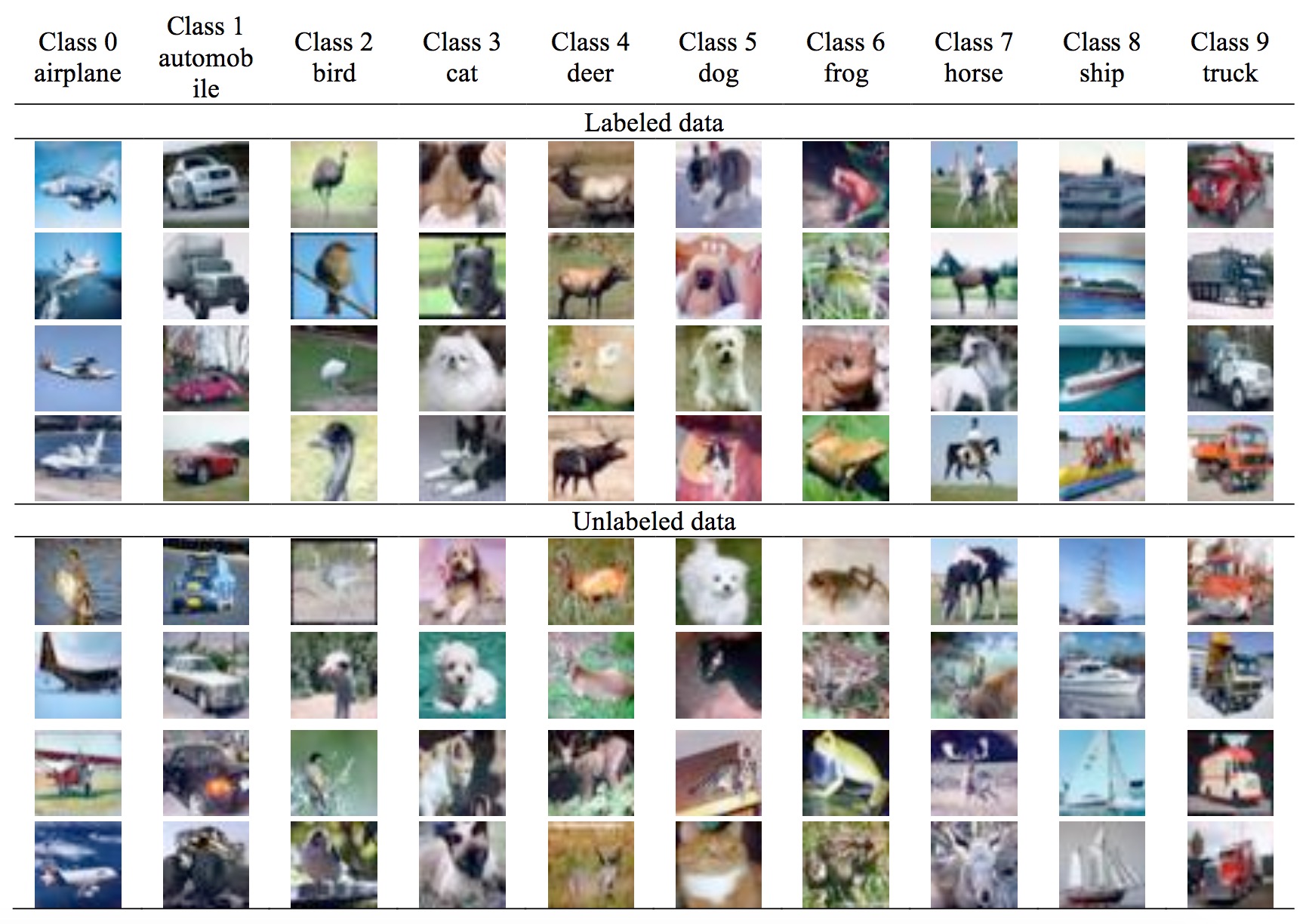}
\end{center}
\vspace{-7pt}
   \caption{Some labeled and unlabeled images of CIFAR-10 after Stage 1.}
\label{fig:visual}
\end{figure*}

\vspace{-7pt}
\paragraph{Qualitative results of Stage 1.}

In Figure~\ref{fig:visual}, we provide some samples of the labeled and unlabeled images of different categories in CIFAR-10. They are the output of the Stage 1 and are the input to the semi-supervised learning approach in Stage 2. We can find that the labeled images are more typical and representative of their corresponding classes than the unlabeled ones. Their foregrounds also stand out more clearly from the background. These verify that Stage 1 works as we expected. It is able to mine the images whose labels are correct and, more importantly, remove the labels of the hard cases whose labels could be not only wrong but difficult to correct.

\begin{table*}
\caption{Some statistics of the correct and incorrect labels by different methods on CIFAR-10 (noisy rate $p= 0.6$)}\label{tab:refine}
\begin{center}
\begin{tabular}{l c c c}
\hline
  & Images with correct labels &  Images with incorrect labels &  Images with no labels \\
\hline\hline
original training set & 70.0\% & 30.0\% & 0.0\% \\
 after~\cite{patrini2016making}'s correction ($\arg\max$) & 54.1\% & 45.9\%&0.0\% \\
 after~\cite{patrini2016making}'s correction (sampling)& 68.1\% & 31.9\%&0.0\% \\   
after Stage 1 of our method & 16.8\% & 0.2\% &83.0\% \\

\hline
\end{tabular}
\end{center}
\vspace{-10pt}
\end{table*}

\vspace{-7pt}
\paragraph{Quantitative results of Stage 1.}
Table~\ref{tab:refine} shows the percentages of the images with correct labels, incorrect labels, and no labels, respectively, of the CIFAR-10 training set before and after Stage 1. We can see that a much smaller percentage (16.75\%) of images are labeled correctly after we apply Stage 1 than the  percentage (70\%) in the original noisy training set. Although this seems like an undesirable situation for the conventional fully supervised methods for learning from noisy labels~\cite{john1995robust,muhlenbach2004identifying}, it actually helps our semi-supervised approach because Stage 1 also reduces the percentage of incorrectly labeled images from 30\% to only 0.2\%, which effectively gets rid of the misleading supervision of the neural network.

For comparison, we have also included the percentages of correct/incorrect labels of~\cite{patrini2016making}. It explicitly infers the label transition matrix (cf.\ Figure~\ref{fig:t}) and uses it to re-weigh the data terms in the loss function. Essentially, it transforms the one-hot labels of an image to a vector of continuous values. We recover the label of the image by either sampling a label from the vector or taking the $\arg\max$ of the vector. The results of both operations are shown in Table~\ref{tab:refine}. Unfortunately, the correction by~\cite{patrini2016making} actually makes the situation worse as more images are incorrectly labeled after the correction. We conjecture that other methods that try to model the label noise could suffer the same issue which then hurt the performance of the neural networks.

\begin{table*}
\caption{Comparison results on CIFAR-10 and MINIST}\label{tab:cifar}
\centering
\begin{tabular}{ l c c c c| c c c c }
\hline
Methods & \multicolumn{4}{c|}{CIFAR-10 14-layer ResNet} & \multicolumn{4}{c}{MNIST fully connected} \\
    & $p=0$ & sy.$p = 0.2$ & asy.$p = 0.2$  & asy.$p = 0.6$ & $p = 0$ & sy.$p = 0.2$ & asy.$p = 0.2$ & asy.$p = 0.6$\\
\hline\hline
cross-entropy~\cite{patrini2016making}& 87.8& 83.7 & 85.0 & 57.6 &  97.9$\pm$ 0.0& 96.9$\pm$ 0.1&97.5$\pm$ 0.0 &53$\pm$ 0.6 \\

unhinged (BN)~\cite{van2015learning}&86.9 &84.1& 83.8& 52.1& 97.6$\pm$ 0.0 &96.9$\pm$ 0.1 &97.0$\pm$ 0.1& 71.2 $\pm$  1.0\\
sigmoid (BN)~\cite{ghosh2015making}&76.0& 66.6&71.8 &57.0&97.2$\pm$ 0.1&93.1$\pm$ 0.1&96.7$\pm$ 0.1&71.4$\pm$ 1.3\\
savage~\cite{masnadi2009design}&80.1 &77.4&76.0& 50.5&97.3$\pm$ 0.0&96.9$\pm$ 0.0&97.0$\pm$ 0.1&51.3$\pm$ 0.4\\
bootstrap soft~\cite{reed2014training}&87.7&84.3&84.6&57.8&97.9$\pm$ 0.0&96.9$\pm$ 0.0&97.5$\pm$ 0.0&53.0$\pm$ 0.4\\
bootstrap hard~\cite{reed2014training}&87.3&83.6&84.7&58.3&97.9$\pm$ 0.0&96.8$\pm$ 0.0&97.4$\pm$ 0.0&55.0$\pm$ 1.3\\

 backward~\cite{patrini2016making} & 87.7 &80.4& 83.8 &66.7 &  97.9$\pm$ 0.0& 96.9$\pm$ 0.0 &96.7$\pm$ 0.1 &67.4$\pm$ 1.5 \\
 forward~\cite{patrini2016making} & 87.4 & 83.4 &\textbf{87.0} & 74.8 & 97.9$\pm$ 0.0 &96.9$\pm$ 0.0&97.7$\pm$ 0.0 &64.9$\pm$ 4.4 \\
\hline
cross-entropy &87.9  &82.4& 85.5 & 56.2 & 98.0$\pm$ 0.1 & 97.1$\pm$ 0.1&97.6$\pm$ 0.2&52.9$\pm$ 0.6 \ \\
improved baseline &87.8 &83.6& 85.2 & 74.1 & 98.0$\pm$ 0.1 &97.1$\pm$ 0.1 &97.7$\pm$ 0.1 &\textbf{76.7$\pm$ 1.6} \\ 
\textbf{ours} &  \textbf{88.0} & \textbf{84.5}& 85.6 & \textbf{75.8} &  \textbf{98.2$\pm$ 0.1} &\textbf{97.7$\pm$ 0.4}  &\textbf{97.8$\pm$ 0.1}  &\textbf{83.4$\pm$ 1.3}  \\
\hline
\end{tabular}
\end{table*}

\vspace{-7pt}
\paragraph{Quantitative results of Stage 2. }
Finally, we apply our semi-supervised method using the processed training set of Stage 1. The corresponding results are shown in Table~\ref{tab:cifar}. In addition to the baseline results reported by~\cite{patrini2016making} (the row of cross-entropy and by our own implementation (the row of cross-entropy, i.e., directly training the neural network using the noisy set), we also include the results of the improved baseline, i.e., balancing different classes and refined validation set using the approach proposed in Stage 1 in the training stage. For the competing method, we include both versions of~\cite{patrini2016making}, which are the best published methods by the time we submit the paper, as far as we know, on the three datasets studied in this paper. However, we exclude the results of the groundtruth transformation matrix in~\cite{patrini2016making} because such a ground truth is usually unknown in practice.

From Table~\ref{tab:cifar}, we can see that our method outperforms the baselines and the existing approach~\cite{patrini2016making} to a large margin when the noisy rate is as high as $p=0.6$. When the noisy rate is smaller ($p=0$ and $p=0.2$), our approach performs better or about the same as~\cite{patrini2016making}. These results verify our modeling hypothesis that, instead of attempting to correct all the noisy labels, the alternative direction can be more effective by ignoring some of the labels in exchange for a small and yet pure labeled set. The more noisy the labels are, the harder to correct the labels or to infer the weights in the loss as done in the existing methods of learning from noisy labels, and the more advantageous our two-stage framework is.


\begin{table*}
\begin{center}
\caption{Comparison results on the Clothing1M dataset~\cite{xiao2015learning}.}\label{tab:cloth}
\begin{tabular}{ c l lll c c}
\hline
   \# &   model &loss / method &initialization & training set& accuracy (reported) & accuracy (our impl.)\\
\hline\hline
1 &  AlexNet & pseudo-label~\cite{lee2013pseudo} & \#9 & 1M, 50K & 73.04 &--\\
2 &  AlexNet & bottom-up~\cite{sukhbaatar2014training} & \#9 & 1M, 50K & 76.22 &--\\
3 &  AlexNet & label noise model~\cite{xiao2015learning} & \#9 & 1M, 50K & {{78.24}} &--\\
4 & 50-ResNet &cross-entropy& ImageNet & 1M & 68.94 &69.03\\
5 & 50-ResNet  &backward~\cite{patrini2016making} & ImageNet& 1M & 69.13 &--\\
6 &50-ResNet  & forward~\cite{patrini2016making} &ImageNet& 1M & {{69.84}}&--\\
7 &  50-ResNet   & \textbf{ours}  & ImageNet& 1M &--& {{77.34}} \\
8 &  50-ResNet   & \textbf{ours}  & ImageNet& 1M, 50K & --&\textbf{79.38} \\
\hline
9 &  AlexNet & cross-entropy & ImageNet & {\textcolor{white}{1M, }}50K & 72.63 &--\\
10 &50-ResNet  & cross-entropy &ImageNet& {\textcolor{white}{1M, }}50K & 75.19 &74.84\\

11 &  50-ResNet   &cross-entropy  & \#6 & {\textcolor{white}{1M, }}50K & 80.38 &--\\
12 &  50-ResNet   &cross-entropy  & \#7  & {\textcolor{white}{1M, }}50K &--& {80.44} \\
13 &  50-ResNet   &cross-entropy  & \#8 & {\textcolor{white}{1M, }}50K & --& \textbf{80.53} \\
\hline
\end{tabular}
\end{center}
\vspace{-10pt}
\end{table*}

\subsection{Experiments on Clothing1M}
The Clothing1M dataset provides a small clean training set of 50K images. The total number of images with noisy labels is about 1M. We experiment both with and without the direct usage of that clean set for the clothing classification task, and reconcile our experiment setup with that used by~\cite{patrini2016making}.

We use the 50-layer ResNet~\cite{he2016deep} pre-trained on ImageNet~\cite{russakovsky2015imagenet} as our base neural network classifier. The balancing cost is $\alpha = 100$. Recall that we overly sample the labeled images in the semi-supervised training process so that each mini-batch has the same numbers of labeled and unlabeled examples. We call every pass over all the labeled images a sub-epoch, and in total we run the experiments for 100 sub-epochs.The ADAM optimizer with a learning rate of 3e-4 is used. We divide the learning rate by half when the validation accuracy does not improve for 10 consecutive sub-epochs. We also apply early stopping when the validation accuracy does not improve for 20 consecutive sub-epochs. 

Table~\ref{tab:cloth} compares our method with several existing ones. We mainly compare to the loss re-weighting method~\cite{patrini2016making} (rows \#5 and \#6), which estimates backward and forward label transition matrices and achieves state-of-the-art results on the Clothing1M. In addition, we also include three earlier methods whose results are reported in~\cite{patrini2016making} on the Clothing1M: a one-stage semi-supervised approach using pseudo labels~\cite{lee2013pseudo} (row \#1), a bottom-up training method for deep neural networks with noisy labels~\cite{sukhbaatar2014training} (row \#2), and a label noise model~\cite{xiao2015learning} (row \#3). Finally, we report the results of the baseline neural networks that are trained using the small clean set (50K) and the large noisy set (1M), respectively, with the vanilla cross-entropy loss (rows \#4, \#9, and \#10). From the table, we can see that our two-stage method significantly outperforms all competing approaches. In particular, our 77.34\% accuracy is much better than the 69.84\% accuracy by the forward label transition~\cite{patrini2016making}. 



Finally, we fine-tune our neural network classifiers (rows \#7 and \#8) over the small clean set with a very small learning rate. It is interesting to see that this increases the accuracy with noticeable margins. The same applies to the neural network obtained in~\cite{patrini2016making} (cf.\ from row \#6 to row \#11). This verifies the necessity of human annotations, and, on the other hand, it also indicates the need of effective algorithms that can learn from the noisy labels. Given a new visual recognition task, it seems like a reasonable strategy to learn from the noisy training set first and then fine-tune the model with a small manually labeled clean set.


\section{Conclusion and discussion}

In this paper, we propose a semi-supervised two-stage approach for learning from noisy labels. We devise two techniques to mine the noisy set in the first stage depending on whether or not there  is clean data available to the learning agent. After that, we train a deep neural network using a semi-supervised learning method.  In our approach, we do not need to know any prior knowledge or estimate any distribution of the noisy labels. 


Our approach outperforms the existing methods especially when the noisy rate is high. This confirms our modeling intuition that the network could be misled by the incorrect labels, which are inevitable by the existing label correction methods for learning from the noisy labels. In contrast, we avoid this explicit effort of label correction by completely ignoring the labels of some data points thanks to the use of the semi-supervised learning strategy. Other realization of our two-stage framework will be explored in the future work.

\vspace{-15pt}

\paragraph{Acknowledgements.} 
This work was supported in part by NSF-1741431.

{\small
\nocite{diersen2011classification, huang2013scalable, guo2011model}
\bibliographystyle{ieee}
\bibliography{egbib}
}

\end{document}